\documentclass[10pt,twocolumn,letterpaper]{article}
\usepackage{authblk}
\usepackage{cvpr}              

\newcommand{\mycolorbox}[1]{
  \raisebox{0.5ex}{\fcolorbox{#1}{#1}{\rule{0ex}{0ex}}} 
}
\usepackage{tcolorbox}
\usepackage{multirow} 
\usepackage{float}
\usepackage{xcolor}
\usepackage{makecell}
\usepackage[accsupp]{axessibility}
\definecolor{deepgreen}{RGB}{0,100,0}  

\newcommand{\reviewer}[1]{%
  \ifnum#1=1
    \textcolor{blue}{\textbf{UEqG}}%
  \else\ifnum#1=2
    \textcolor{red}{\textbf{epbW}}%
  \else\ifnum#1=3
    \textcolor{deepgreen}{\textbf{Jw1y}}%
  \fi\fi\fi
}

\definecolor{lightyellow}{RGB}{255, 255, 204}
\definecolor{lightgreen}{RGB}{243, 249, 238}
\definecolor{lightred}{RGB}{255, 229, 229}

\definecolor{cvprblue}{rgb}{0.21,0.49,0.74}
\definecolor{seagreen}{rgb}{0.34,0.55,0.2}
\usepackage[pagebackref,breaklinks,colorlinks,allcolors=cvprblue]{hyperref}

\def\paperID{7355} 
\def\confName{CVPR}
\def\confYear{2025}

\title{MPDrive: Improving Spatial Understanding with Marker-Based Prompt Learning for Autonomous Driving}

\author{ Zhiyuan Zhang\textsuperscript{1,$*$}, Xiaofan Li\textsuperscript{2,$*$}, Zhihao Xu\textsuperscript{1,$*$}, Wenjie Peng\textsuperscript{1},\\ Zijian Zhou\textsuperscript{3}, Miaojing Shi\textsuperscript{4}, Shuangping Huang\textsuperscript{1,5,$\dagger$}}

\affil{%
  \textsuperscript{1}South China University of Technology, 
  \textsuperscript{2}Baidu Inc., \\
  \textsuperscript{3}King's College London, 
  \textsuperscript{4}Tongji University,
  \textsuperscript{5}Pazhou Laboratory
}

\begin{document}
\maketitle
\footnotetext[1]{These authors contribute equally.}
\footnotetext[2]{Corresponding Author.}
\begin{abstract}
Autonomous driving visual question answering (AD-VQA) aims to answer questions related to perception, prediction, and planning based on given driving scene images, heavily relying on the model's spatial understanding capabilities.
Prior works typically express spatial information through textual representations of coordinates, resulting in semantic gaps between visual coordinate representations and textual descriptions.
This oversight hinders the accurate transmission of spatial information and increases the expressive burden.
To address this, we propose a novel \textbf{M}arker-based \textbf{P}rompt learning framework \textbf{(MPDrive)}, which represents spatial coordinates by concise visual markers, ensuring linguistic expressive consistency and enhancing the accuracy of both visual perception and spatial expression in AD-VQA.
Specifically, we create marker images by employing a detection expert to overlay object regions with numerical labels, converting complex textual coordinate generation into straightforward text-based visual marker predictions.
Moreover, we fuse original and marker images as scene-level features and integrate them with detection priors to derive instance-level features. By combining these features, we construct dual-granularity visual prompts that stimulate the LLM’s spatial perception capabilities.
Extensive experiments on the DriveLM and CODA-LM datasets show that MPDrive achieves state-of-the-art performance, particularly in cases requiring sophisticated spatial understanding.
\end{abstract}

\section{Introduction}
\label{sec:intro}

\begin{figure}[tbp]
    \centering
    \includegraphics[width=0.93\linewidth]{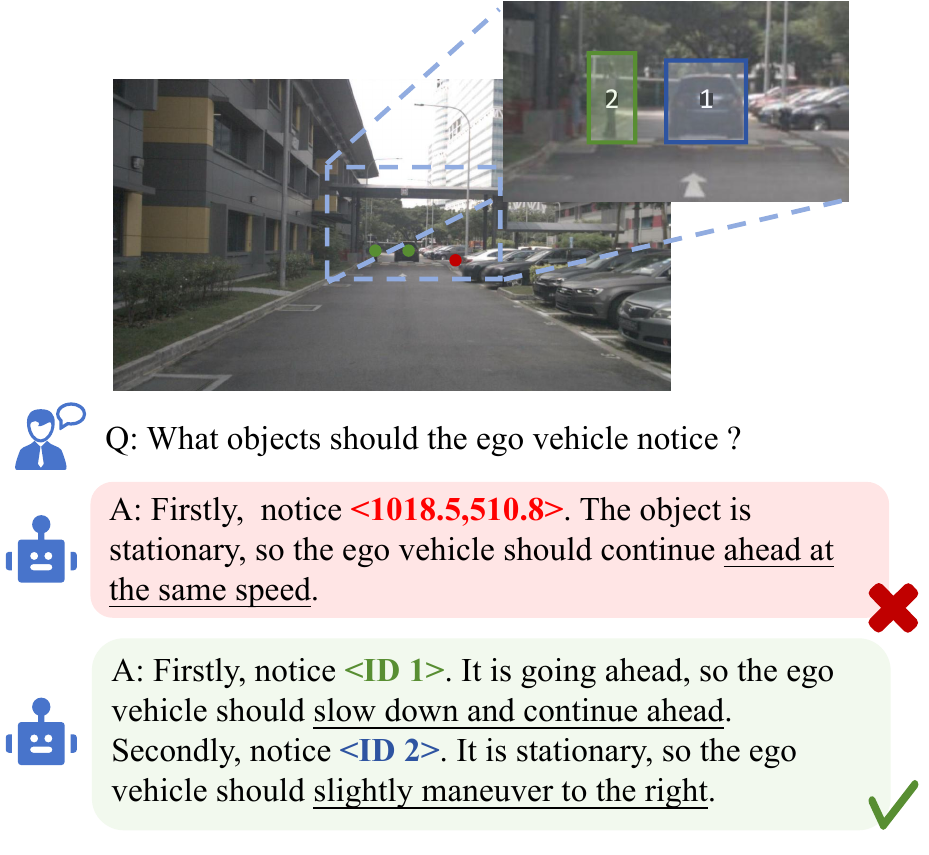}
    \caption{Comparison of object response process between mainstream MLLMs (\textcolor{red}{red box}) and our proposed MPDrive (\textcolor{seagreen}{green box}). Current research directly represents object spatial coordinates in text format, leading to semantic gaps between coordinates and text descriptions. This misalignment adversely impacts subsequent prediction and planning tasks. In contrast, MPDrive converts complex spatial coordinate generation into text-based visual marker (region with numerical label) predictions, ensuring linguistic consistency. }
    \label{fig:illstrate}
\end{figure}

Autonomous driving has advanced rapidly, showing potential to enhance road safety, traffic efficiency, and reduce human error~\citep{bevformer, bevformer2, languageprompt, wen2023dilu}.
A robust autonomous driving system requires an agent capable of perceiving complex environments and making informed decisions.
Recently, Multi-modal Large Language Models (MLLMs) have emerged as a promising approach for autonomous driving, demonstrating strong generalization capabilities in visual question answering (AD-VQA) tasks~\citep{drivegpt4, copa, lmmwm, zhou2024embodied,drivewithllms, Dolphins, NuScenesqa, Talk2BEV, TOKEN}.

Current MLLMs face challenges in spatial understanding for autonomous driving scenarios~\citep{DriveVLM, coda-lm, MiniGPT4}, limiting their ability to accurately locate, identify, and describe objects and their states in driving scenes.
While several AD-VQA methods~\citep{reason2drive, nuScenesMQA, coda-lm, drivelm, DRAMA} have attempted to enhance MLLM performance through instruction tuning on domain-specific datasets, they have not adequately addressed the core challenge of spatial reasoning optimization. 
Among these approaches, some methods~\citep{reason2drive, DriveVLM} enhance spatial understanding by integrating detection priors.
However, these methods typically express spatial coordinates in textual format, leading to inconsistencies between coordinate-based and linguistic descriptions~\citep{som,pix2seq,PIVOT}, which undermines the perceptual accuracy and precise spatial expression in autonomous driving.

In this paper, we focus on enhancing the consistency of coordinate representations and spatial understanding in autonomous driving.  
We propose \textbf{M}arker-Based \textbf{P}rompt Learning (\textbf{MPDrive}), a novel multi-modal framework that uses text indices to annotate each traffic element and directly predicts the coordinates of the corresponding index. 
As shown in Figure \ref{fig:illstrate}, MPDrive utilizes visual markers, implemented as text-based indices overlaid on detected regions in the images, to highlight the spatial location of key objects. This transformation simplifies the complex process of spatial coordinate generation into a text-based visual marker prediction, thereby bridging the gap between coordinate representations and linguistic descriptions in AD-VQA.
Additionally, by incorporating multi-level spatial features, MPDrive stimulates LLM's spatial perception capabilities to enhance the accuracy of visual marker prediction, boosting performance in predictions and planning tasks.

To this end, we propose two components: the Marker ControlNet (MCNet) and the Perception-Enhanced Spatial Prompt Learning (PSPL).
Specifically, MCNet processes both the original and visual marker images, accurately expressing spatial information while preserving original image features.
PSPL combines scene- and instance-level visual prompts:
i) MCNet generates scene-level prompts to capture comprehensive spatial relations, while ii) instance-level prompts incorporate fine-grained object features through masked average pooling.
This integration significantly enhances MPDrive's spatial understanding capabilities.


In summary, our contributions are as follows:
\begin{itemize}
    \item We propose MPDrive, a \textbf{M}arker-based \textbf{P}rompt learning framework that leverages visual markers to bridge the gap between coordinate-based and linguistic descriptions in AD-VQA, significantly improving the spatial understanding in autonomous driving.
    \item MPDrive consists of two components: the Marker ControlNet (MCNet) and the Perception-Enhanced Spatial Prompt Learning (PSPL).
    MCNet fuses visual marker images for scene features, while PSPL integrates scene- and instance-level visual prompts to enhance multi-level spatial understanding.
    \item Extensive experiments demonstrate that MPDrive achieves state-of-the-art results on AD-VQA tasks, excelling on multi-image tasks with the DriveLM dataset~\citep{drivelm} and single-image tasks with the CODA-LM dataset~\citep{coda-lm}, particularly in complex spatial scenarios.
\end{itemize}

\section{Related Work}
\label{sec:relate}

\subsection{AD-VQA}
AD-VQA has emerged as an essential component for promoting human-vehicle interaction and improving decision-making in complex driving scenarios~\citep{LLM4Drive}. Recent autonomous driving research has advanced through multiple perspectives: multi-modal fusion for scene understanding~\citep{NuScenesqa}, multi-step reasoning for decision-making~\citep{drivelm, DriveCoT}, signal control optimization~\citep{DME-Driver}, motion planning~\citep{GPT-Driver}, and corner-case handling~\citep{coda-lm}. These approaches collectively enhance the system's capabilities through effective integration of multi-modal data and reasoning mechanisms.

Recent research has increasingly focused on enhancing the spatial understanding capabilities of MLLMs in autonomous driving. ELM~\citep{zhou2024embodied} leverages expert-generated textual descriptions to improve object localization, while LLM-Driver~\citep{drivewithllms} advances context understanding by integrating vectorized numeric modalities with pre-trained LLMs. Similarly, Reason2Drive~\citep{reason2drive} employs a prior tokenizer and an instructed vision decoder to strengthen visual localization capabilities. 
Although these strategies aim to enhance spatial understanding through detection priors, they often involve complex training schemes, such as the addition of intricate network architectures or detection optimization functions. Furthermore, these strategies typically represent spatial coordinates in text format, which may increase the complexity of the model. Consequently, these approaches neglect the discrepancies between coordinate-based and linguistic descriptions, which compromises perceptual accuracy and the precise articulation of spatial information in autonomous driving systems.

\begin{figure*}[tbp]
    \centering
    \includegraphics[width=0.92\linewidth]{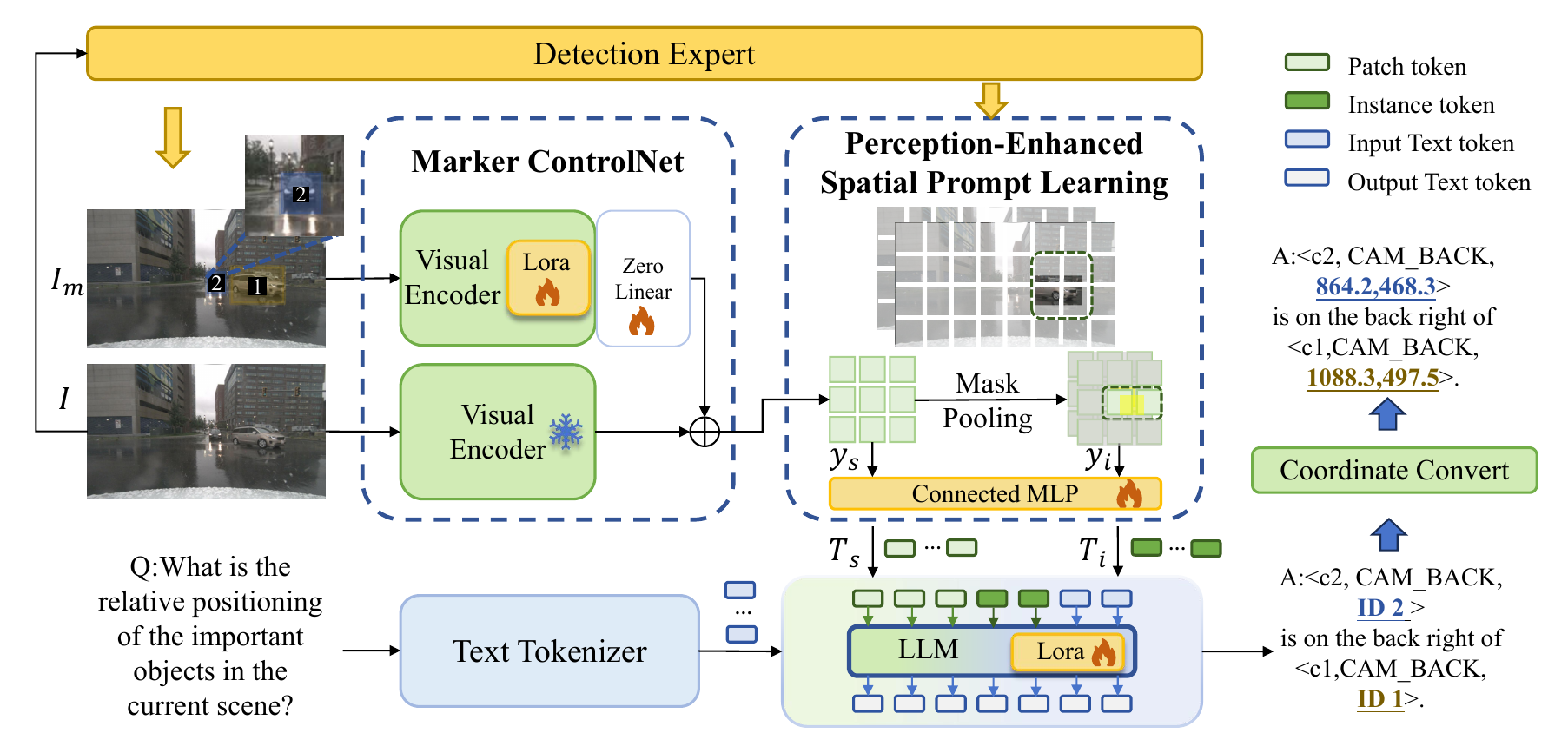}
    \caption{Overview of the MPDrive framework. For clarity, we illustrate the process using a single-view image. The detection expert generates a visual marker image $I_m$.
    This marker image $I_m$ and the original image $I$ are processed by MCNet to extract scene-level features $y_s$.
    For the Perception-Enhanced Spatial Prompt Learning module, these scene-level features $y_s$ undergo mask average pooling for each instance mask to obtain instance-level features $y_i$. 
    Subsequently, both scene-level features $y_s$ and instance-level features $y_i$ are processed through a connected MLP to generate visual prompts $T_s$ and $T_i$ respectively. 
    Finally, these visual prompts, combined with text embeddings, are fed into the Large Language Model to generate the output $\hat{s}$. For coordinate prediction, MPDrive predicts the marker index $k$  corresponding to the target object and then converts it into the respective coordinates.}
    \label{fig:method}
\end{figure*}

\subsection{MLLMs}

MLLMs have demonstrated remarkable interpretability and generalization capabilities~\citep{PIVOT, copa, Pink,}. Recent advances in MLLMs primarily focus on vision-language alignment and training strategies. For alignment, BLIP-2~\citep{blip-2} introduces Q-Former for efficient modality bridging, MiniGPT-4~\citep{MiniGPT4} aligns frozen visual encoders with LLMs through projection layers, and InternVL~\citep{internvl} proposes progressive alignment between vision models and LLMs. For training strategies, LLAVA~\citep{llava} utilizes machine-generated instruction data, while MiniCPM~\citep{mincpm} optimizes performance through advanced learning rate scheduling. These advances have enabled MLLMs' successful applications in video understanding~\citep{loscv, MoReVQA}, image understanding~\citep{VITRON, LLM-CXR,dai2023disentangling,dai2024one,peng2024globally}, and embodied AI~\citep{PaLM-e,Socratic}.

In autonomous driving, MLLMs have been explored in various ways. 
Atlas~\citep{altas} and DriveGPT4~\citep{drivegpt4} enhance driving capabilities through 3D tokenization and multi-frame video processing, respectively. For resource efficiency, MiniDrive~\citep{minidrive} and EM-VLM4AD~\citep{vlm4ad} provide lightweight MLLMs for autonomous driving. Meanwhile, TOKEN~\citep{TOKEN} integrates tokenized object-level knowledge, while DriveAdapter~\citep{DriveAdapter} improves model performance through feature alignment and action-guided learning.
These efforts attempt to apply MLLMs in autonomous driving; however, they have not sufficiently explored spatial understanding in driving scenarios.

\subsection{Visual Prompts}

Visual prompts have been extensively used for transfer and adaptation across various downstream tasks~\citep{vpt,cpl,lpd,gvp} and can be categorized into learnable and image-modifying approaches. Learnable visual prompt methods incorporate trainable tokens as additional visual inputs~\citep{mplm,vpt,mvlpt}, with works like LM-BFF~\citep{mplm} and VPT~\citep{vpt} demonstrating enhanced learning efficiency through prompt-based fine-tuning. Image-modifying visual prompt methods focus on altering images with expert-generated elements~\citep{som,fgvp,api}, where FGVP~\citep{fgvp}, API~\citep{api}, and SoM~\citep{som} have shown significant improvements in MLLMs' visual understanding through techniques such as segmentation masks and attention heatmaps.

While our approach draws inspiration from SoM~\citep{som}, which overlays masks and markers on images, we have introduced several key improvements to better address the specific challenges in autonomous driving tasks.
First, conventional markers may obscure critical information in the original image, such as object colors and features. To address this, we employ Marker ControlNet to introduce visual markers that gradually incorporate marker-derived information, thereby preserving key visual information from the original image while leveraging the benefits of visual markers. 
Additionally, we incorporate a visual prompt process: Perception-Enhanced Spatial Prompt Learning, which includes scene-level and instance-level visual prompts, significantly enhancing the spatial perception capabilities of MPDrive.

\section{Method}
\label{sec:method}

This section is organized as follows. Section~\ref{sec:mtd_pre} introduces the preliminary knowledge, including the task definition and MLLM pipline. Section~\ref{sec:mtd_vm} details the Visual Marker. Section~\ref{sec:mtd_arch} presents two core modules of MPDrive: Marker ControlNet (MCNet) and Perception-Enhanced Spatial Prompt Learning (PSPL).

\subsection{Preliminary}
\label{sec:mtd_pre}
Given a set of $m$ view images $\{ I_1, I_2, \ldots, I_m \}$ and a text question $Q$, AD-VQA aims to generate a response sequence $\hat{S} = ({\hat{s}}_1, {\hat{s}}_2, \ldots, {\hat{s}}_N)$, where ${\hat{s}}_i$ denotes the $i$-th token in the sequence of length $N$.
The workflow of MLLMs in AD-VQA is as follows: 1) a visual encoder that extracts visual features from each view $I_i$; 2) a connected MLP that transforms multi-view features into image tokens; 3) a text tokenizer that converts the question $Q$ into text tokens; and 4) an LLM that fuses image tokens and text tokens to generate the response sequence ${\hat{S}}$.


Building upon these MLLMs, we propose MPDrive to enhance spatial understanding capabilities.
For clarity, we illustrate the methodology using a single-view scenario, while noting that all operations naturally extend to multi-view cases.

\subsection{Visual Marker}
\label{sec:mtd_vm}

To bridge the gap between spatial coordinate representations and linguistic descriptions, we introduce Visual Marker. This approach simplifies the task of spatial coordinate generation by converting it into straightforward text-based visual marker predictions. As illustrated in Figure \ref{fig:method}, given an input image $I \in \mathbb{R}^{H \times W \times 3}$, we use a detection expert, StreamPETR \cite{streamP}, to identify traffic objects~(\eg{}, cars, trucks, and buses), following the object categories specified in~\citep{streamP}.
The detection expert generates $K$ object masks, represented as binary masks $R = [r_1, r_2, \ldots, r_K]$, where $r_k \in \{0, 1\}^{H \times W}$ denotes the $k$-th detection mask. 
For $r_k$, we compute its average centroid coordinates $c_k=(x_{k},y_{k})$, which represents the central location of this object. 
The annotated marker image $I_m$ is generated by modifying the original image $I$ through two steps: First, annotating the marker index $k$ at each object's centroid $c_k=(x_{k},y_{k})$, and second, overlaying corresponding semi-transparent mask regions $r_k$ to describe object boundaries.
Furthermore, when new spatial coordinates $c_{new}$~(more than $d_{th}$ pixels from existing coordinates) are referenced in the question $Q$, we assign them a marker index $K+1$ and annotate the index on $I_m$ to maintain consistent spatial reasoning across visual and text modalities.

For response generation, we leverage the Visual Marker to improve the effectiveness of visual prompts and ensure consistency of language output. Specifically, the LLM first generates the indicator $k$ from the given images and the question, then maps this index $k$ to its corresponding centroid coordinates $c_k = (x_k, y_k)$ for precise localization.
This process allows MPDrive to identify the key objects by their markers, while complex spatial perception is handled by the Detection Expert. By avoiding direct coordinate output, this approach mitigates the linguistic complexity for LLMs, ensuring consistent text output.

\subsection{MPDrive Architecture}
\label{sec:mtd_arch}

As illustrated in Figure \ref{fig:method}, MPDrive consists of two key components: MCNet and PSPL.
MCNet enhances spatial representation by leveraging both the original image and the additional visual marker image to achieve dual-level fused scene features. 
Based on these extracted features and the detection expert, PSPL generates scene-level and instance-level visual prompts, thereby enhancing the understanding of driving scene information and object information. The integration of these components significantly boosts the spatial perception capabilities of MPDrive.

\noindent\textbf{Marker ControlNet.}
To effectively preserve key features of the original image and fully leverage the rich information in visual markers, we propose the Marker ControlNet (MCNet). 
This module takes both the original image and the visual marker image as input, and generates scene-level features.

We freeze the parameters $ \theta$ of the original visual encoder $E$ and create a trainable copy with parameters $\theta_c$, denoted as $E_c$. During training, the original visual encoder remains frozen, and we focus on training the new control block using Low-Rank Adaptation (LoRA) \cite{lora} on the multi-head attention modules and the feed-forward networks with the rank of 16. We connect the original visual encoder and the control block with a zero linear, $Z$, where both weight and bias are initialized to zero, with parameters $\theta_z$. These layers are trained alongside the control block, allowing for effective parameter tuning and improved performance.
The original image features are extracted using the original visual encoder $E$, while the visual marker image features are extracted using the new control block, $E_c$ combined with $Z$. These features are combined through element-wise addition for scene-level feature fusion:

\begin{equation}
    y_s = E(I;\theta) + Z(E_c(I_m;\theta_c);\theta_z),
    \label{eq:control}
\end{equation}
where $y_s$ represents the scene-level features.

Since the weight and bias parameters of the zero linear layer are initialized to zero, the $Z$ term in Equation \ref{eq:control} starts with zero, thereby preserving the integrity of the original image features. During subsequent optimization phases, beneficial features from the visual marker image will be gradually introduced through backpropagation.

MCNet effectively incorporates visual markers, enabling MPDrive to learn additional semantic information through the guidance of visual markers while preserving the critical features of the original image. More importantly, this approach ensures that MPDrive can capture the visual marker information and then output the corresponding text-based markers, thereby maintaining consistency in linguistic output when generating spatial information.

\noindent\textbf{Perception-Enhanced Spatial Prompt Learning.}
To address the limitations of MLLMs in spatial expression capabilities, we introduce Perception-Enhanced Spatial Prompt Learning (PSPL), aiming at enhancing the spatial perception of MPDrive by utilizing both scene-level and instance-level visual prompts. 

Visual markers in images accurately represent the spatial information for the entire scene. Therefore, the output features $y_s$ of MCNet encompass rich scene-level spatial information.
Subsequently, $y_s$ is processed through the connected MLP to generate scene-level visual prompts  $T_s$.
These scene-level visual prompts significantly improve the perception and accurate understanding of spatial information in complex scenarios.

To further enhance the representation of spatial information at the instance level, we introduce instance-level visual prompts. 
Given the $k$-th detection object with its region mask $r_k$, the scene-level visual prompts $y_s \in \mathbb{R}^{H' \times W' \times C}$, where $C$ is the number of channels, $W'$ is the width, and $H'$ is the height, we resize the binary region mask $r_k$ into the same size as $y_s$ and use mask average pooling:

\begin{equation}
    y^k_{i} = MAP(y_s,r_k),
\end{equation}
where $MAP$ represents the mask average pooling operation, and $y^k_{i}$ denotes the $k$-th instance-level visual features.

Given $K$ objects, we obtain a set of instance-level visual features $\{y^1_{i},\ldots,y^K_{i}\}$. These features are processed through a Connected MLP to generate instance-level visual prompts $T_{i}$.
This instance-level visual prompts enriches object spatial representation. PSPL concatenates the scene-level visual prompts $T_s$ and the instance-level visual prompts $T_i$ together, enhancing the spatial perception ability of MPDrive. 

\noindent\textbf{Large Language Model.}
The LLM receives input text tokens from the text tokenizer and spatial prompts $T_s$ and $T_i$ from the PSPL module. It processes these inputs using its internal model, where LoRA is applied to both multi-head attention modules and feed-forward networks at a rank of 16, generating an output sequence $\hat{S}=(\hat{s}_1, \hat{s}_2, \ldots, \hat{s}_N)$ of $N$ words. The output token sequence $\hat{S}$ is then used to compute cross-entropy loss with the ground truth sequence $S=(s_1, s_2, \ldots, s_N)$:

\begin{equation}
    Loss = -\sum_{i=1}^{n} s_i \log(\hat{s}_i).
\end{equation}

\section{Experiments}
\label{sec:exp}
\begin{table*}[ht]
\centering
\begin{tabular}{l|c|c|ccccc}
\hline
\multirow{2}{*}{Method} & \multirow{2}{*}{\begin{tabular}[c]{@{}c@{}}Inference\\ Schema\end{tabular}} & \begin{tabular}[c]{@{}c@{}}Spatial$\uparrow$\\ Perception\end{tabular} & \multicolumn{5}{c}{{Language$\uparrow$}}                                                                          \\ \cline{3-8} 
                                 &                                                                                      & {Match}                                                        & {Accuracy} & {BLEU-4} & {ROUGE\_L} & {CIDEr} & {METEOR}                    \\ \hline
DriveLM-Agent~\citep{drivelm}                  & Graph                                                                                & -                                                                     & -                 & \textbf{53.09}            & 66.79             & 2.79           & 36.19                              \\ \hline
EM-VLM4AD~\citep{vlm4ad}                      & Single                                                                               & -                                                                     & -                 & 45.36            & 71.98             & 3.20           & 34.49                              \\
MiniDrive~\citep{minidrive}                      & Single                                                                               & -                                                                     & -                 & 50.20            & 73.50             & 3.32           & \underline{37.40}                              \\
LLaMA-Adapter~\citep{llama-adapter} & Single & 1.48              & 66.66             & 45.96             & 69.78             & 3.07             & 33.66 \\
InternVL-2~\citep{internvl}                  & Single                                                                               & \underline{7.59 }                                                                 & \underline{82.54}             & 51.42            & \textbf{77.08}    & \underline{3.53}           & 37.12                              \\
\textbf{Ours: MPDrive}                             & Single                                                                               & \textbf{13.43}                                                        & \textbf{85.18}    & \underline{52.71 }  & \underline{76.98}             & \textbf{3.56}  & \textbf{38.31}\\ \hline
\end{tabular}
\caption {Quantitative evaluation on the DriveLM dataset. MPDrive significantly outperforms existing approaches in both spatial perception and language understanding metrics. \textbf{Bold} indicates the highest value, while an \underline{underline} indicates the second-highest value.}
\label{tbl:result_drive}
\end{table*}

\begin{table*}[ht]
\centering
\begin{tabular}{l|c|cccccc|c}
\hline
\multirow{2}{*}{{Method}} & \multirow{2}{*}{\begin{tabular}[c]{@{}c@{}}{General$\uparrow$}\\ {Text-Score}\end{tabular}} & \multicolumn{6}{c|}{{Regional Perception$\uparrow$}}                               & \multirow{2}{*}{\begin{tabular}[c]{@{}c@{}}{Suggestion$\uparrow$}\\ { Text-Score}\end{tabular}} \\ \cline{3-8}
                        &                                                                               & \multicolumn{1}{c|}{{ALL}}   & {Vehicle} & {VRU}   & {Cone}  & {Barrier} & {Other} &                                                                                  \\ \hline
LLaVA1.5~\citep{llava}             & 22.60                                                                         & \multicolumn{1}{c|}{34.78} & 40.00   & 28.00 & 32.22 & 24.00   & 10.00 & 14.20                                                                            \\
Qwen-VL-Chat~\citep{qwen}          & 26.00                                                                         & \multicolumn{1}{c|}{53.33} & 57.76   & \underline{60.00} & 48.89 & 44.29   & 35.71 & 35.40                                                                            \\
Qwen-VL-Max~\citep{qwen}            & \underline{34.60 }                                                                        & \multicolumn{1}{c|}{\underline{68.17}} & \underline{69.83}   & 56.00 & \underline{80.00} & 59.29   & \textbf{65.71} & \underline{47.40}                                                                            \\
MiniDrive~\citep{minidrive}        & 24.60                                                                         & \multicolumn{1}{c|}{66.34} & 67.41   & 36.00 & \textbf{84.44} & \underline{62.86}   & \underline{62.85} & 45.44                                                                            \\
\textbf{Ours: MPDrive}                     & \textbf{41.80}                                                                            & \multicolumn{1}{c|}{\textbf{76.12}}      & \textbf{79.48}      & \textbf{70.00}    &  77.77    &  \textbf{70.00} & \underline{62.85}      & \textbf{58.20}    \\ \hline
\end{tabular}
\caption {Quantitative evaluation on the CODA-LM dataset. MPDrive achieves superior performance across all evaluation metrics.  \textbf{Bold} indicates the highest value, while an \underline{underline} indicates the second-hightest value.}
\label{tbl:result_coda}
\end{table*}

In this section, we comprehensively evaluate the efficacy of MPDrive. Section~\ref{sec:exp_es} introduces the experimental setup, while Section~\ref{sec:exp_quan} and Section~\ref{sec:exp_qual} provide an in-depth analysis of the quantitative and qualitative results. Lastly, Section~\ref{sec:exp_as} presents ablation studies to evaluate the contribution of each component.

\subsection{Experimental Setting}
\label{sec:exp_es}

\textbf{Datasets.} 
We conduct experiments on the DriveLM~\citep{drivelm} and CODA-LM~\citep{coda-lm} datasets. For the DriveLM dataset, we follow the data partitioning strategy employed by EM-VLM4AD~\citep{vlm4ad} and MiniDrive~\citep{minidrive}, which divides the dataset into training and validation subsets, allocating 70\% and 30\% of the data, respectively. The training set comprises 341,353 unique QA pairs, while the validation set contains 18,817 distinct QA pairs. Each QA pair consists of six view images: front view, left front view, right front view, back view, left back view, and right back view. For the CODA-LM dataset, we train MPDrive using a training set of 20,495 QA pairs and validate it with a mini set of 193 QA pairs. Each QA pair consists of a front-view image.

\noindent\textbf{Evaluation Metrics.} 
To facilitate a rigorous and fair comparison, we adopt the evaluation metrics consistent with those used in EM-VLM4AD and MiniDrive studies, including the BLEU-4~\citep{bleu}, ROUGE\_L~\citep{ROUGE}, CIDEr~\citep{CIDEr}, and METEOR~\citep{METEOR}. 
These metrics evaluate the linguistic consistency between predicted values and ground truth through overlap, recall, consensus-based evaluation, and semantic similarity, reflecting the perception, prediction, and planning capabilities of MLLMs.
Additionally, following the CVPR 2024 Autonomous Driving Challenge guidelines~\citep{zhou2024embodied,drivelm}, we incorporate additional performance metrics: match and accuracy. The match metric quantifies the percentage of predicted center point coordinates that have an Euclidean distance of less than 16 pixels from the ground truth, providing an intuitive validation of the spatial information expression capabilities of MLLMs. Accuracy evaluates the correctness of responses in multiple-choice and yes/no questions, offering a comprehensive assessment of MLLMs' capabilities.

\noindent\textbf{Implementation Details.} 
\label{subsec:imp}
During the training phase, we employ a cosine learning schedule with an initial rate of $5e-4$ and utilize the AdamW~\citep{adamw} optimizer with a weight decay of $0.01$. For the DriveLM dataset, we employ a batch size of 128 and conducted training for 3,000 iterations across eight A800 GPUs, equivalent to approximately 1 epoch. For the CODA-LM dataset, we conducted training for 2000 iterations, equivalent to approximately 12 epochs. Throughout the entire training process, the visual encoder weights remained frozen. We fine-tune the connected MLP and zero MLP while applying Low-Rank Adaptation (LoRA)~\citep{lora} to both the visual encoder within MCNet and the LLM decoder. For both the training and inference stages, we resize the input image resolution to $448\times448$ pixels. The number of detected objects $K$ is dynamically determined by the detection expert for each image, with a maximum limit of 100 objects across all camera views. We set new spatial coordinates $d_{th}=50$.

\subsection{Quantitative Results}
\label{sec:exp_quan}

\begin{figure*}[tbp]
    \centering
    \includegraphics[width=0.95\linewidth]{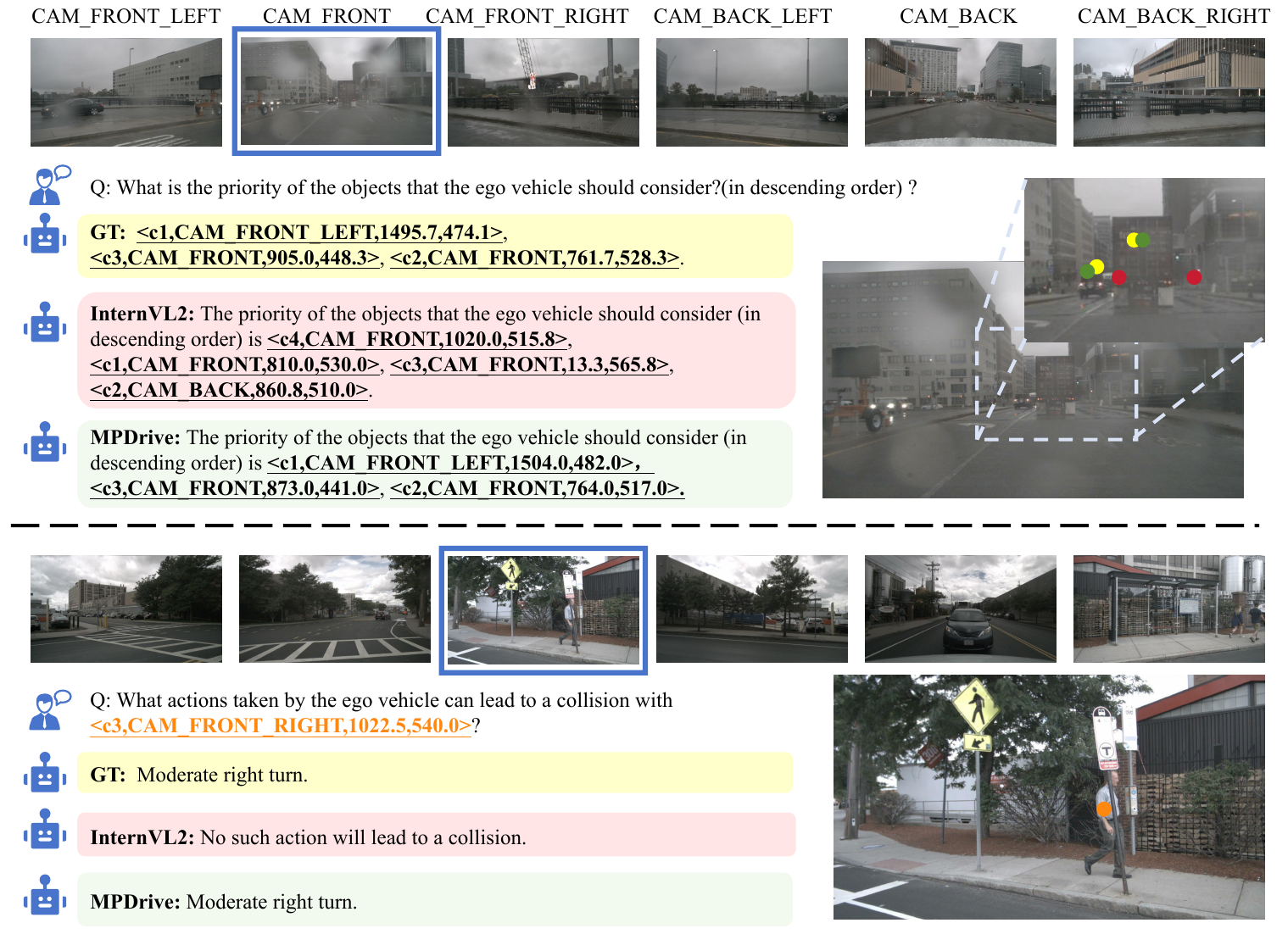}
    \caption{Comparison of the responses between InternVL-2 and our proposed MPDrive. The yellow (\mycolorbox{yellow}) area and dots represent the response and coordinates of ground truth (GT), the green (\mycolorbox{green}) area and dots indicate the response and coordinates of MPDrive, the red (\mycolorbox{red}) area and dots denote the response and coordinates of InternVL-2. The blue box ($\color{blue}\boxed{\phantom{a}}$) indicates the image that is most relevant to the response, with an enlarged version of this image located in the bottom right corner of each sample, the orange dots (\textcolor{orange}{$\bullet$}) represent the positions of the coordinates in the image related to the question. }
    \label{fig:qual}
\end{figure*}

We conduct the quantitative evaluation with competitive methods on the DriveLM dataset to demonstrate the effectiveness of MPDrive, as shown in Table \ref{tbl:result_drive}. Our proposed method demonstrates outstanding performance, particularly in the CIDEr and METEOR metrics, achieving scores of 3.56 and 38.31, respectively. Furthermore, it outperforms all single-turn inference approaches in BLEU-4, closely approximating the performance of the graph-based multi-turn reasoning method (DriveLM-Agent), indicating its superior performance in linguistic consistency.
Additionally, MPDrive demonstrates strong spatial perception abilities with a match score of 13.43 and an accuracy of 85.18, which surpasses the performance of InternVL-2.

As shown in Table~\ref{tbl:result_coda}, MPDrive demonstrates remarkable performance across various tasks on the CODA-LM dataset.
In the general perception task, MPDrive achieves a score of 41.80, significantly outperforming other competitive methods. This indicates its superior ability to perceive and interpret driving scenes effectively.
For the spatially relevant region perception task, MPDrive excells in several subcategories. It achieves a score of 79.48 in the vehicle category and 70.00 in the VRU~(Vulnerable Road Users) category, underscoring its fine-grained perception capabilities for spatial objects. Additionally, it performs well in the cone (77.77), barrier (70.00), and other (62.85) categories, highlighting its comprehensive spatial understanding.
MPDrive achieves a top score of 58.20 in driving suggestion generation, demonstrating superior spatial awareness and planning capabilities for effective driving recommendations.

These results validate MPDrive for precise spatial expression and demonstrate MPDrive's enhanced spatial perception capabilities in autonomous driving scenarios.

\subsection{Qualitative Examples}
\label{sec:exp_qual}

In Figure \ref{fig:qual}, we compare the actual response results of MPDrive with InternVL-2 on unseen samples, evaluating the spatial perception and the task planning capabilities of MPDrive.
In the upper sample of Figure \ref{fig:qual}, we display the predicted coordinates from one of the most relevant images. The predictions of InternVL-2 are located in incorrect areas, while MPDrive locates the important objects, aligning with ground truth annotations. This demonstrates superior spatial understanding capabilities of MPDrive.

In the lower example of Figure \ref{fig:qual}, when asked to identify dangerous behaviors involving vehicles and pedestrians, InternVL-2 incorrectly concludes no collision risk with the pedestrian. In contrast, MPDrive accurately assesses the vehicle-pedestrian spatial relationship, leading to correct planning decisions. This demonstrates the advanced ability of MPDrive to analyze complex scenarios and make precise decisions, highlighting its effectiveness in autonomous driving applications.
More qualitative examples can be found in the supplementary materials.

In conclusion, MPDrive outperforms InternVL-2 on unseen samples, exhibiting accurate object localization and reliable assessment of spatial relations, which are crucial for safe autonomous driving.

\subsection{Ablation Studies}
\label{sec:exp_as}

\begin{table*}[t]
\centering
\begin{tabular}{cc|c|c|ccccc}
\hline
\multicolumn{2}{c|}{\multirow{2}{*}{\begin{tabular}[c]{@{}c@{}}Scene-level\\ Visual Prompts\end{tabular}}} & \multirow{3}{*}{\begin{tabular}[c]{@{}c@{}}Instance-level\\ Visual Prompts\end{tabular}} & \multirow{2}{*}{\begin{tabular}[c]{@{}c@{}}Spatial$\uparrow$ \\ Perception\end{tabular}} & \multicolumn{5}{c}{\multirow{2}{*}{Language$\uparrow$}}  \\
\multicolumn{2}{c|}{}                                                                                     &                                                                                         &                                                                               & \multicolumn{5}{c}{}                           \\ \cline{1-2} \cline{4-9} 
\multicolumn{1}{c|}{Visual Marker}                                 & MCNet                                &                                                                                         & Match                                                                         & Accuracy & BLEU-4 & ROUGE\_L & CIDEr  & METEOR \\ \hline
\multicolumn{1}{c|}{-}                                             & -                                    & -                                                                                       & 7.59                                                                          & \underline{82.54 }   & 51.42  & \textbf{77.08}    & \underline{3.53} & 37.12  \\
\multicolumn{1}{c|}{\checkmark}                                             & -                                    & -                                                                                       & \underline{11.89}                                                                         & 80.42    & 52.04  & 76.17    & 3.50   & 37.88  \\
\multicolumn{1}{c|}{\checkmark}                                             & \checkmark                                    & -                                                                                       & 9.70                                                                          & 78.83    & \underline{52.56}  & 76.53    & \underline{3.53}   & \underline{38.14}  \\
\multicolumn{1}{c|}{\checkmark}                                             & \checkmark                                    & \checkmark                                                                                       & \textbf{13.43}                                                                       & \textbf{85.18}    & \textbf{52.71}  & \underline{76.98}    & \textbf{3.56} & \textbf{38.31}  \\ \hline
\end{tabular}
\caption{Ablation experiments on different parts of MPDrvie on the DriveLM dataset.  \textbf{Bold} indicates the highest value, while an \underline{underline} indicates the second-hightest value. }
\label{tbl:ablation}
\end{table*}

\begin{table*}[ht]
\centering
\begin{tabular}{l|c|c|ccccc}
\hline
\multirow{2}{*}{Method} & \multirow{2}{*}{\makecell[c]{MPDrive}} & \makecell[c]{Spatial\\ Perception}$\uparrow$ & \multicolumn{5}{c}{{Language~$\uparrow$}} \\
\cline{3-8}
 & & {Match} & {Accuracy} & {BLEU-4} & {ROUGE\_L} & {CIDEr} & {METEOR} \\
\hline
\multirow{2}{*}{LLaMA-Adapter~\citep{llama-adapter}} & - & 1.48              & 66.66             & 45.96             & 69.78             & 3.07             & 33.66 \\
& \checkmark & \underline{10.05} & 68.25             & 47.97             & 73.54             & 3.28             & 35.58 \\
\hline
\multirow{2}{*}{InternVL-2~\citep{internvl}} & - & 7.59              & \underline{82.54} & \underline{51.42} & \textbf{77.08}    & \underline{3.53} & \underline{37.12} \\
& \checkmark & \textbf{13.43}    & \textbf{85.18}    & \textbf{52.71} & \underline{76.98} & \textbf{3.56}    & \textbf{38.31}\\
\hline
\end{tabular}
\caption {Ablation studies of MPDrive using different MLLMs on the DriveLM dataset.
MPDrive significantly enhances the spatial understanding performance of MLLMs. 
 \textbf{Bold} indicates the highest value, while an \underline{underline} indicates the second-highest value.}
\label{tbl:result_llm}
\end{table*}

In this section, we conduct ablation studies on visual markers, MCNet, and instance-level visual prompts. Furthermore, we evaluate MPDrive across different MLLMs. To ensure a fair comparison, we conducted ablation experiments on the DriveLM dataset~\citep{drivelm}, which includes six-view images and encompasses perception, prediction, and planning tasks, thereby facilitating a comprehensive assessment of MLLM in autonomous driving scenarios. Furthermore, the various evaluation metrics on the DriveLM dataset can evaluate the performance of MPDrive from multiple perspectives.

\noindent\textbf{Scene-level Visual Prompts.}
To evaluate the effectiveness of the scene-level visual prompts, we conduct ablation experiments for Visual Marker and MCNet. 
Table \ref{tbl:ablation} presents the ablation study of scene-level prompts. Visual marker significantly improves spatial perception, as seen in the match score from 7.59 to 11.89. However, its impact on language metrics shows mixed results. While the accuracy slightly decreases to 80.42, the improvements in BLEU-4 and METEOR scores indicate enhanced linguistic expression consistency in MPDrive. We attribute this performance to the potential feature interference between Visual Markers and object features in the visual space.

By incorporating the MCNet, most metrics for measuring language consistency have improved. While the match score decreases from 11.89 to 9.70 compared to using Visual Marker alone, the model achieves better linguistic quality with improved BLEU-4 (52.56) and METEOR (38.14) scores. This suggests that MCNet helps balance the feature representation between spatial information and semantic understanding, though at the cost of some spatial perception capability.

\noindent\textbf{Instance-level Visual Prompts.}
To evaluate the effectiveness of the instance-level visual prompts, we conduct comparative experiments with and without this component while keeping all other settings identical, as shown in Table~\ref{tbl:ablation}.
The integration of instance-level visual prompts leads to comprehensive improvements across both spatial and language metrics. Specifically, the match score further increases to 13.43, surpassing all previous configurations, while the accuracy achieves the highest value of 85.18. Moreover, the language generation quality consistently improves, with BLEU-4 reaching 52.71, ROUGE\_L at 76.98, CIDEr at 3.56, and METEOR achieving 38.31. These results demonstrate that the instance-level visual prompts effectively enhance both spatial perception and language understanding, suggesting its crucial role in precise text-based marker index prediction.

\noindent\textbf{Different MLLMs.}
To assess the model-agnostic nature of MPDrive, we extend our experiments to include LLaMA-Adapter as an alternative MLLM. Table~\ref{tbl:result_llm} demonstrates that applying our MPDrive framework to LLaMA-Adapter yields significant performance gains compared to the original LLaMA-Adapter implementation.
Specifically, MPDrive~(LLaMA-Adapter) achieves a significantly higher match score of 10.05 compared to LLaMA-Adapter’s 1.48, indicating a substantial enhancement in spatial perception capabilities. In terms of language generation metrics, MPDrive~(LLaMA-Adapter) outperforms LLaMA-Adapter in all aspects: BLEU-4 increases from 45.96 to 47.97, ROUGE-L improves from 69.78 to 73.54, CIDEr rises from 3.07 to 3.28, and METEOR advances from 33.66 to 35.58. Additionally, MPDrive demonstrates slightly higher Accuracy at 68.25 compared to 66.66. The comparative analysis indicates that MPDrive effectively enhances the spatial understanding of different MLLMs.

\section{Conclusion}
\label{sec:con}

We introduce a novel MLLM-based framework called MPDrive for AD-VQA. 
MPDrive transforms complex spatial coordinate generation into concise visual marker predictions. It incorporates MCNet and PSPL to enhance both scene-level and instance-level spatial perception capabilities. MPDrive achieves state-of-the-art performance on the multi-view input autonomous driving task with the DriveLM dataset, as well as on the single-view input task with the CODA-LM dataset.

MPDrive relies on a prior expert for spatial perception and language expression, and errors from the expert can affect its performance. Furthermore, although MPDrive enhances the spatial perception capabilities of AD-VQA, the long-horizon temporal perception remains a significant challenge in autonomous driving. Therefore, exploring how to advance this research based on MPDrive is worthy of further investigation.


{
    \small
    \bibliographystyle{ieeenat_fullname}
    \bibliography{main}
}

\clearpage
\setcounter{page}{1}
\maketitlesupplementary

\setcounter{section}{0}

In this supplementary material, we provide additional details regarding MPDrive and present further ablation studies. Initially, we define the evaluation metrics used to assess performance on the DriveLM dataset. These metrics include accuracy and matching scores.
Subsequently, we conducted a series of ablation studies to investigate the effects of different detection experts and image token lengths.
For qualitative analysis, we present three types of examples. Firstly, we provide visual prompt comparisons to highlight the differences between MPDrive and InternVL-2. Secondly, ablation examples are included to demonstrate the impact of each component on the generated responses. Finally, we present comparative case studies showcasing the performance differences between MPDrive and InternVL-2.

\begin{table*}[ht]
\centering
\begin{tabular}{c|c|ccccc}
\hline
\multirow{2}{*}{Method}  & \begin{tabular}[c]{@{}c@{}}Spatial$\uparrow$\\ Perception\end{tabular} & \multicolumn{5}{c}{{Language$\uparrow$}}                                                                          \\ \cline{2-7} 
 & {Match}                                                        & {Accuracy} & {BLEU-4} & {ROUGE\_L} & {CIDEr} & {METEOR}                    \\  \hline
MPDrvie~(64)                             & \textbf{13.76}                                                       & 79.37    & 52.35   & 76.95            & 3.54  & 38.10\\ 
MPDrive~(256)                             & 13.43                                                       & \textbf{85.18}    & \textbf{52.71}   & \textbf{76.98 }            & \textbf{3.56}  & \textbf{38.31}\\ \hline
\end{tabular}
\caption {Ablation study of different image tokens. }
\label{tbl:result_token}
\end{table*}

\begin{table*}[ht]
\centering
\begin{tabular}{c|c|c|ccccc}
\hline
\multirow{2}{*}{Method} & \multirow{2}{*}{\begin{tabular}[c]{@{}c@{}}mAP\end{tabular}} & \begin{tabular}[c]{@{}c@{}}Spatial$\uparrow$\\ Perception\end{tabular} & \multicolumn{5}{c}{{Language$\uparrow$}}                                                                          \\ \cline{3-8} 
                                 &                                                                                      & {Match}                                                        & {Accuracy} & {BLEU-4} & {ROUGE\_L} & {CIDEr} & {METEOR}                    \\ \hline
MPDrvie~(DETR3D)   &       \textbf{ 50.10 }                & \textbf{13.76}                                                       & 83.30    & 52.40   & \textbf{76.99 }           & \textbf{3.58}  & 37.38\\ 
MPDrvie~(StreamPetr)   &      48.20                   & 13.43                                                       &\textbf{ 85.18}    & \textbf{52.71}   & 76.98             & 3.56  & \textbf{38.31}\\ \hline
\end{tabular}
\caption {Ablation study of different detection experts. }
\label{tbl:result_expert}
\end{table*}

\section{More Evaluation Details}

\paragraph{Accuracy Metric}
For the DriveLM dataset, both multi-choice questions and yes/no questions are used to calculate the accuracy score.
The multi-choice questions include perception and behavior prediction. For perception questions, the question is ``What is the moving status of the object?''. We will provide 7 candidate options, randomly selecting 3 options from the incorrect answers and incorporating the correct answer to construct the multiple-choice question. Similarly, for behavior prediction questions, the question is ``Predict the behavior of the ego vehicle.'', with a total of 21 candidate options.
The yes/no questions include perception, prediction, and planning, and the ground truth annotations only contain ``yes'' or ``no''.

Given $m$ predicted responses $\hat{S}=(\hat{r}_1,\hat{r}_2,...,\hat{r}_m)$ and the ground truth answers $R=(r_1,r_2,...,r_m)$, the accuracy score can be calculated as follows:

\begin{equation}
     Acc = \sum_{i=1}^{m} {\frac{\hat{r}_i==r_i}{m}},
\end{equation}
where ${\hat{r}_i==r_i}$ is a boolean expression: it equals 1 if the predicted response matches the ground truth, and 0 otherwise.

\section{Ablation Study on Different Image Token Lengths}

\paragraph{Match Metric}
For the DriveLM dataset, we extracted $l_{gt}$ center coordinates $P=[p_1, p_2, ..., p_{l_{gt}}]$ from the ground truth responses and $l_{p}$ center coordinates $\hat{P}=[\hat{p}_1, \hat{p}_2, ...,\hat{p}_{l_{p}}]$ from the predicted responses. We then calculated the proportion of coordinates in the predicted responses that have an Euclidean distance of less than 16 from the ground truth coordinates, thus obtaining the matching ratio, formulated as:

\begin{equation}
     Match =  {\frac{min(\left|\left| P-\hat{P} \right|\right|_2)<16}{l_{gt}}},
\end{equation}
where ${min(\left|\left| P-\hat{P} \right|\right|_2)<16}$ represents the number of pairs of points between the $P$ and $\hat{P}$ for which the minimum Euclidean distance is less than 16 among all possible matches.

\section{Ablation Study on Different Detection Experts}

To investigate the impact of detection expert performance on spatial localization accuracy, we conducted a comparative analysis using two distinct detection models: StreamPetr and DETR3D, which achieve mAP scores of $48.20$ and $50.10$, respectively, on the NuScenes Val Set, as shown in Table~\ref{tbl:result_expert}. Experimental results indicate a positive correlation between detector performance and spatial localization accuracy. Higher-performing detectors generally exhibit improved spatial localization.

To examine the effect of different image token lengths, we experiment with compressing scene-level tokens from 256 to 64 per image, thereby reducing the total scene tokens from 1,536 to 384 for six view images. 
As shown in table~\ref{tbl:result_token}, this token compression strategy led to a degradation in model performance on the DriveLM dataset. Specifically, the decline in accuracy metrics suggests that reducing the number of image tokens compromised the model's ability to effectively capture and process visual information.

\section{Qualitative Visual Prompt Comparison}

\begin{figure}[tbp]
    \centering
    \includegraphics[width=\linewidth]{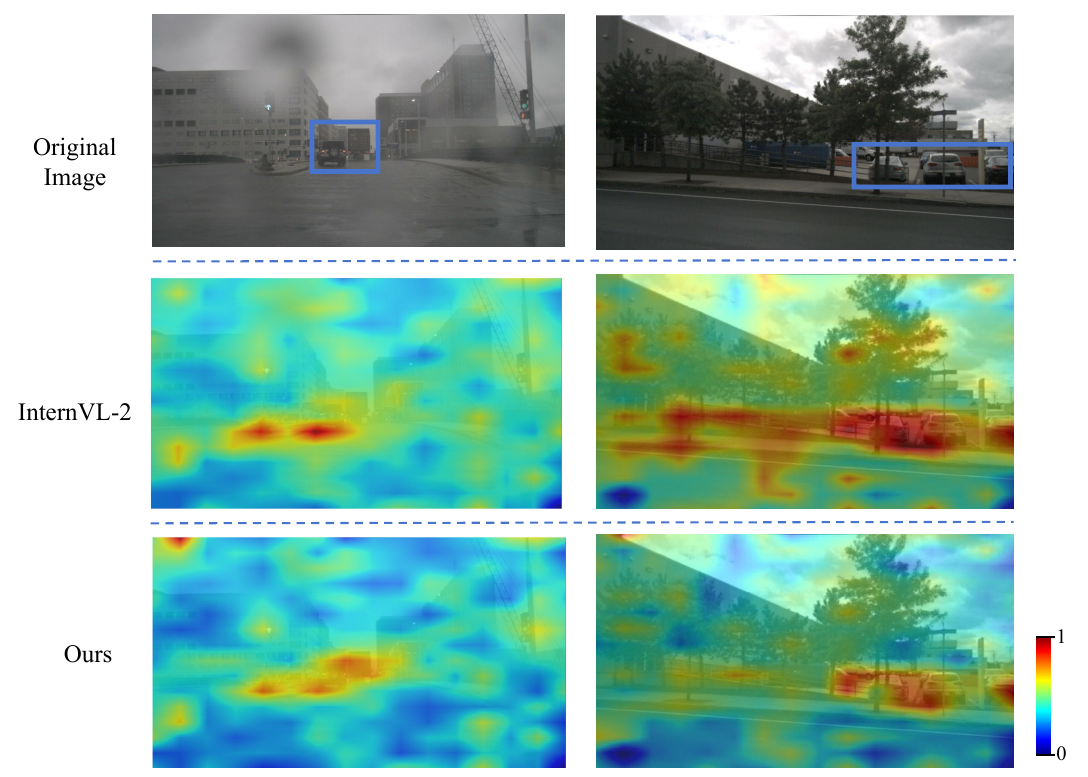}
    \caption{Visual prompt activation examples between InternVL-2 and our proposed MPDrive. }
    \label{fig:supp_vp}
\end{figure}

To demonstrate the effectiveness of our spatial-enhanced features, we conduct a qualitative analysis comparing the visual prompt activation maps between MPDrive and InternVL-2 (Figure \ref{fig:supp_vp}). While both models exhibit scene awareness for common road elements (e.g., trucks, cars, and traffic signs), InternVL-2 shows notable limitations in object perception, either failing to detect key objects (as shown in the left example where it misses the truck on the road) or displaying redundant activation in non-informative road areas (as illustrated in the right example). In contrast, MPDrive demonstrates more focused and efficient spatial attention by activating only task-relevant vehicles, eliminating redundant information, and maintaining better spatial localization of relevant objects.

\section{Qualitative Ablation Examples}

\begin{figure*}[hb]
    \centering
    \includegraphics[width=0.95\linewidth]{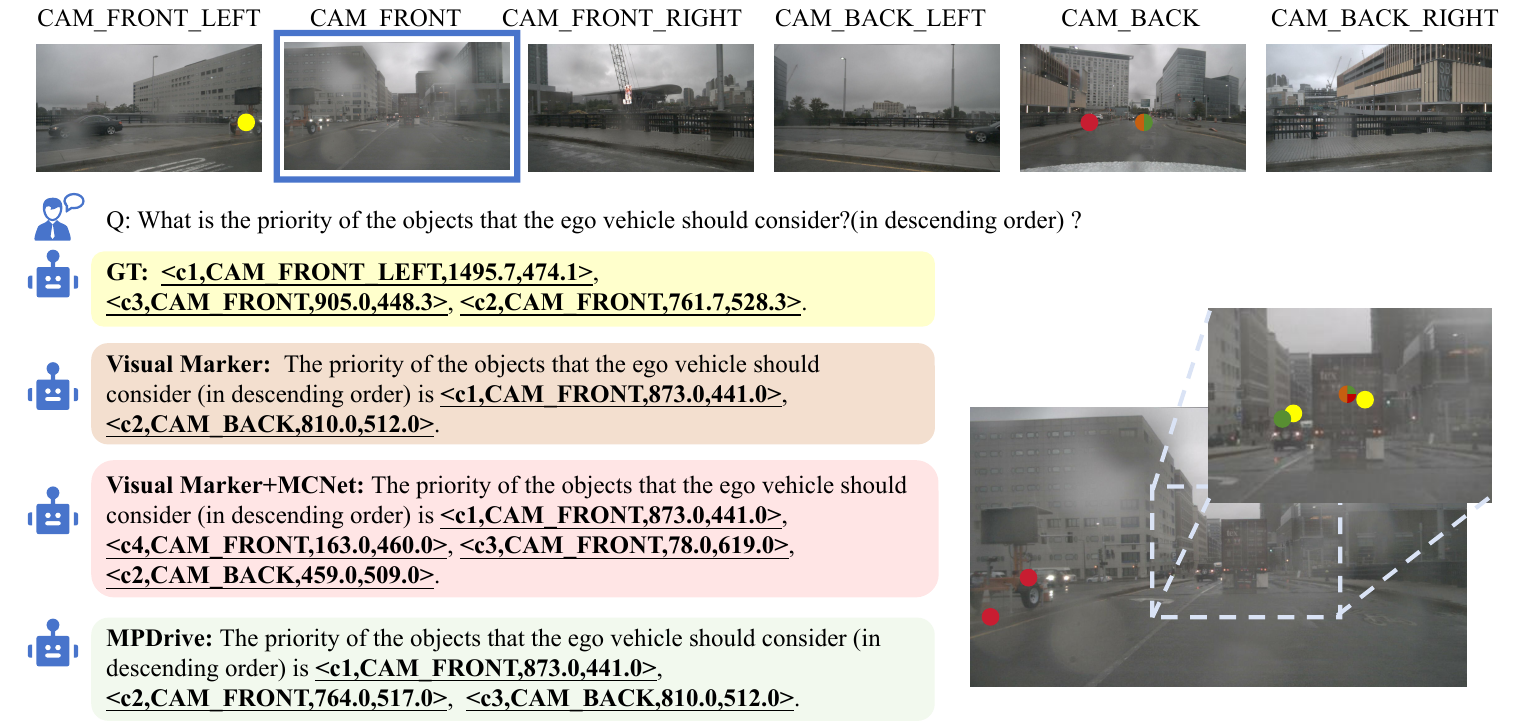}
    \caption{Comparison of different components of MPDrive on the responses. The yellow (\mycolorbox{yellow}) area and dots represent the response and coordinates of ground truth (GT), the brown (\mycolorbox{brown}) area and dots indicate the response and coordinates after adding the Visual Marker, the red (\mycolorbox{red}) area and dots denote the response and coordinates after adding the Visual Marker and the MCNet, and the green (\mycolorbox{green}) area and dots indicate the response and coordinates of MPDrive.}
    \label{fig:supp_ab}
\end{figure*}

Figure \ref{fig:supp_ab} demonstrates the impact of different components of MPDrive on the responses, we display the predicted coordinates from one of the most relevant images, and after introducing the Visual Marker, the predicted coordinates contain one correct answer. Following the incorporation of MCNet, the model output multiple coordinates in the front-view image, all of which were located on objects; however, the answer included irrelevant objects such as barriers and trucks. With the addition of the instance-level visual prompt, the model was able to locate each coordinate accurately. This sample indicates that the Visual Marker and MCNet contribute to the precise representation of the spatial coordinates of objects, ensuring consistency in language expression. Meanwhile, the instance-level prompt enhances the spatial features of the objects, further improving the spatial perception capabilities of MPDrive.

\section{More Qualitative Examples}
\begin{figure*}[tbp]
    \centering
    \includegraphics[width=0.95\linewidth]{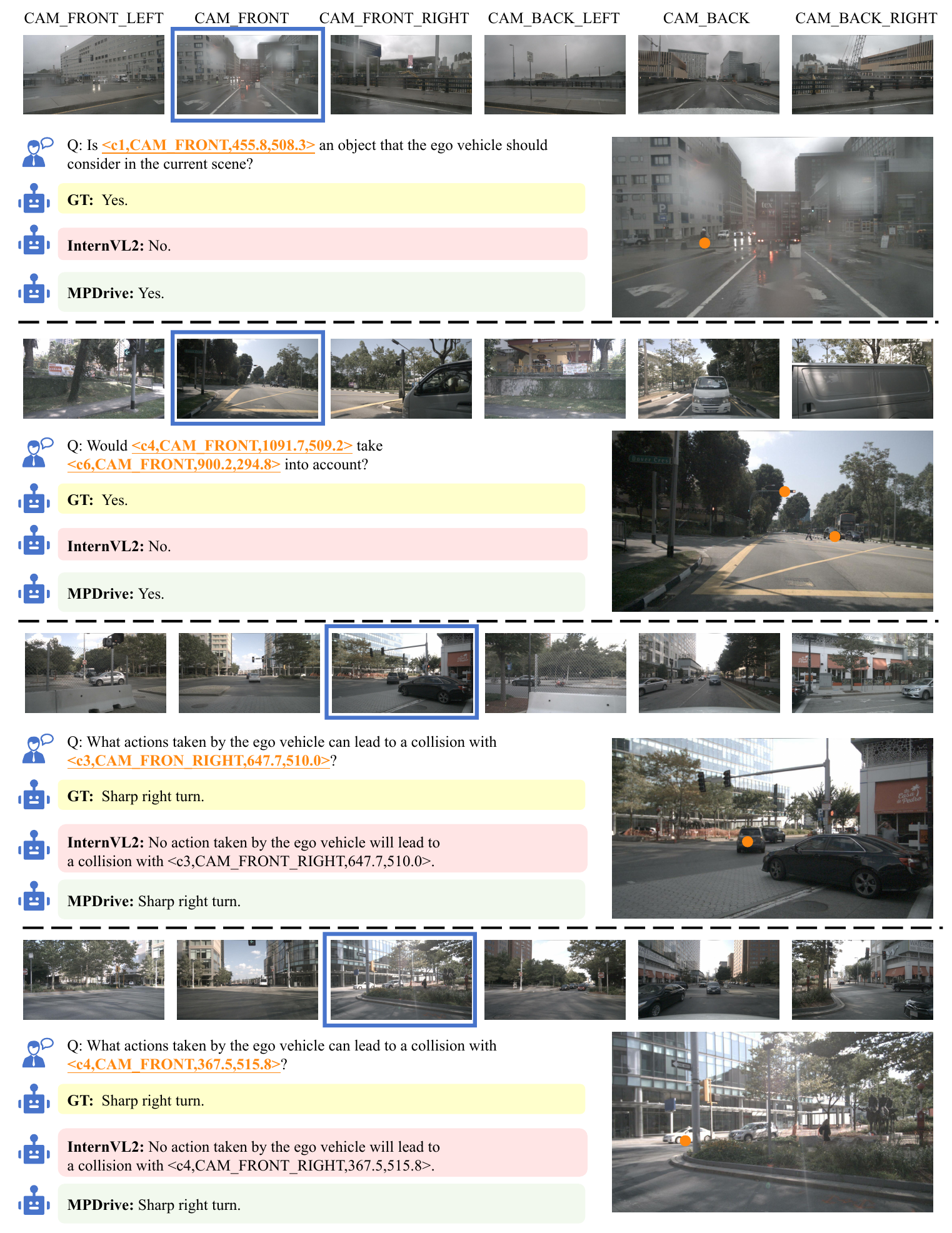}
    \caption{Comparison of the responses between InternVL-2 and our proposed MPDrive. }
    \label{fig:supp}
\end{figure*}

In this section, we present more qualitative examples of MPDrive responses. Figure \ref{fig:supp} illustrates a comparison between the response results of MPDrive and InternVL-2. 
In the first sample of Figure \ref{fig:supp}, for the question of identifying whether the mentioned pedestrian is an important object, InternVL-2 incorrectly answers that the pedestrian crossing the street is not an object that should be considered, however, the pedestrian on the left side is indeed significant because the ego vehicle is making a left turn, and MPDrive provides an accurate assessment in this scenario.
Similarly, in the second sample, for the question of understanding the relationship between the mentioned vehicle and the traffic light, InternVL-2 incorrectly assumes that the car is unrelated to the traffic light. However, the traffic light signals influence the vehicle's position. MPDrive, with its excellent spatial perception abilities, can accurately recognize the relationship between the car and the traffic light.
In the last two samples, for the questions of identifying the dangerous behaviors between the ego vehicle and other vehicles, InternVL-2 struggles to recognize the relative spatial relationships between the ego vehicle and the relevant vehicles due to a lack of strong spatial perception capabilities, thereby limiting its ability to identify potential dangerous behaviors accurately.
In contrast, MPDrive successfully perceives the spatial positions of the relevant vehicles, because of its superior spatial perception abilities, allowing it to make accurate planning decisions.

In summary, MPDrive demonstrates an advantage in scenarios requiring precise spatial perception. Its ability to accurately interpret spatial relationships and identify critical objects allows it to make more informed and safer planning decisions. This enhanced spatial understanding is crucial for the effective navigation and safety of autonomous systems, highlighting the potential of MPDrive for superior performance in complex driving environments.

\end{document}